*Article*

# LiDAR Point Cloud Colourisation Using Multi-Camera Fusion and Low-Light Image Enhancement

Pasindu Ranasinghe [1], Dibyayan Patra [1], Bikram Banerjee [2] and Simit Raval [1,*]

[1] School of Minerals and Energy Resources Engineering, University of New South Wales, Sydney, NSW 2052, Australia; pasindu.ranasinghe@unsw.edu.au (P.R.); d.patra@unsw.edu.au (D.P.); simit.raval@unsw.edu.au (S.R.)
[2] School of Surveying and Built Environment, University of Southern Queensland, Toowoomba, QLD 4350, Australia; bikram.banerjee@unisq.edu.au (B.B.)
[*] Correspondence: simit@unsw.edu.au; Tel.: +61 433 663 423

**Highlights**

**What are the main findings?**

- Developed an end-to-end pipeline combining multi-camera fusion with automated, targetless LiDAR–camera calibration to achieve full 360° real-time colourisation.
- Added colour correction and low-light enhancement modules that recover scene details at illumination as low as 0.5 lx, comparable to well-lit conditions.

**What is the implication of the main finding?**

- Provides a hardware-agnostic solution for reliable monitoring and mapping in low-light environments, including underground mines and night-time navigation.
- Simplifies deployment and improves LiDAR interpretability for applications such as autonomous navigation, geological surveys, and vegetation analysis.

**Abstract**

In recent years, the fusion of camera data with LiDAR measurements has emerged as a powerful approach to enhance spatial understanding. This study introduces a novel, hardware-agnostic methodology that generates colourised point clouds from mechanical LiDAR using multiple camera inputs, providing complete 360-degree coverage. The primary innovation lies in its robustness under low-light conditions, achieved through the integration of a low-light image enhancement module within the fusion pipeline. The system requires initial calibration to determine intrinsic camera parameters, followed by automatic computation of the geometric transformation between the LiDAR and cameras—removing the need for specialised calibration targets and streamlining the setup. The data processing framework uses colour correction to ensure uniformity across camera feeds before fusion. The algorithm was tested using a Velodyne Puck Hi-Res LiDAR and a four-camera configuration. The optimised software achieved real-time performance and reliable colourisation even under very low illumination, successfully recovering scene details that would otherwise remain undetectable.

## 1. Introduction

LiDAR (Light Detection and Ranging) is becoming increasingly popular across various fields due to its versatility and effectiveness. It operates by emitting short pulses of laser light and measuring the time it takes for the light to reflect off a surface and return to the sensor. This technology is one of the key components in the autonomous vehicle industry, used for navigation and object detection. It is widely used in mapping, surveying, and virtual reality. Beyond these applications, LiDAR also plays a crucial role in geological surveys, atmospheric research, and precision agricultural practices. A key advantage of LiDAR over other optical sensors is its ability to produce accurate distance measurements regardless of lighting conditions over long ranges. Despite these strengths, its monochromatic nature makes it challenging for humans and algorithms to interpret scenes or recognise objects, limiting its usefulness in applications requiring detailed semantic information, such as urban mapping, material classification, and vegetation analysis. In autonomous navigation, the lack of colour hinders the identification of road signs, lane markings, and traffic signals. This limitation extends to remote sensing applications such as vegetation classification and terrain analysis, which rely on spectral cues to distinguish land cover types and identify surface composition. Without colour data, these tasks often require additional sensors or more complex processing steps, increasing system complexity and computational cost [1-5].

Cameras, on the other hand, excel in capturing detailed images with rich spectral details and high visual quality. However, a single camera cannot accurately gauge distance information. If depth information is required, it must be inferred through computational methods such as structure-from-motion or deep learning-based depth estimation models. These approaches are often unreliable and inconsistent, especially under low-light conditions. Additionally, they increase processing complexity and computational demands, making them less efficient for real-time applications. Given these complementary strengths and limitations, combining LiDAR and cameras offers a more robust solution: LiDAR contributes accurate depth information, while cameras add colour and texture, together improving spatial awareness, object detection, and segmentation performance [4-6].

To address the lack of spectral detail in LiDAR data, research has focused on methods to colourise point clouds. Two main approaches have emerged: camera-based methods and neural network models. The camera-based approach involves aligning photographic imagery with LiDAR data, which requires the identification of a transformation matrix to align the coordinate frame of these two sensors [1,7-10]. Alternatively, neural network models predict RGB values for each point in the point cloud, offering an automated solution [11-14]. However, neural networks may not accurately replicate true colours due to variations in training data, biases in the learning process, and limited information about environmental conditions. Recent studies further show that such inaccuracies can significantly degrade downstream tasks such as semantic segmentation, in some cases performing worse than geometry-only inputs [15].

Given these limitations, camera-based methods are often preferred for their ability to produce more reliable and realistic colourised point clouds. Xu et al. introduced a novel colourisation algorithm that operates without a position and orientation system using a registration approach based on normalised Zernike moments [2]. Similarly, Pavel Vechersky et al. presented a technique that uses independent cameras, synchronising them by aligning the calculated yaw rates of the camera and LiDAR [1]. Another notable contribution comes from Ting and Chan et al., who outline an efficient colourisation method using point-to-pixel orthogonal projection [9].

Central to all these colourisation methods is the accurate alignment between LiDAR and camera coordinate frames. To achieve this, numerous studies have focused on determining the transformation matrix that maps LiDAR points to corresponding camera image pixels. A common approach involves estimating the transformation by detecting planar and linear features of known objects from LiDAR data and visual data from the camera. These corresponding features are then aligned using optimisation algorithms, with the Iterative Closest Point (ICP) being one of the most widely used methods [16-19]. Grammatikopoulos et al. simplify this process by proposing an algorithm using AprilTag fiducial markers and LiDAR reflective targets attached to a planar board [20]. Pandey et al. present an algorithm for automatically calibrating LiDAR and camera systems, bypassing the need for traditional target objects in calibration [21]. Calibration without using known objects eliminates the effort required to create objects and set up the scene. Koide et al. proposed a method for such calibration; however, this approach is limited to specific LiDAR setups, particularly those capable of generating dense point clouds [22]. More recently, learning-based approaches have emerged as a new trend, delivering fully automatic and targetless calibration pipelines. Leahy and Jabari proposed Galibr, a method that uses ground-plane initialisation followed by edge matching to achieve calibration in unstructured environments [23]. Zhou et al. presented DF-Calib, which formulates calibration as a depth-flow estimation problem and reports sub-centimetre translation accuracy and rotation error below $0.05°$ [24]. Zang et al. introduced TLC-Calib, jointly optimising sensor poses and scene geometry through anchored 3D Gaussians and photometric losses [25].

While these advances have improved automation, they continue to share common limitations. All of these existing methods rely on a single-camera setup, leaving more than 75% of the LiDAR scan uncoloured due to the limited field of view provided by a single camera. This incomplete coverage reduces semantic interpretability and limits usefulness in downstream tasks. In addition, many approaches require time-consuming calibration procedures with fiducial targets or multi-view data, increasing setup time and operational complexity. As a result, these methods are not scalable: each new deployment requires significant manual effort and repeated calibration, limiting their practicality in large-scale or real-time applications. Learning-based calibration methods, though promising, depend on large, curated datasets and lengthy training, making them difficult to adapt and poorly generalisable across diverse environments. Another major limitation is their reliance on well-lit environments; under low-light or nighttime conditions, insufficient visual information results in point clouds with poor colour fidelity and incomplete coverage, which limits their accuracy and usefulness in practical applications. These shortcomings restrict the applicability of such systems in challenging settings, including underground mapping, night-time monitoring, and autonomous navigation, where reliable performance under reduced visibility is essential [26].

The method proposed in this study addresses the limitations of prior approaches through a multi-camera setup that achieves complete 360-degree coverage of the LiDAR field of view. The novelty of this work lies in three aspects: (i) the development of a simple, scalable, and generalisable calibration package that eliminates the need for specialised calibration objects or repeated viewpoint captures, while providing fast and reliable performance (ii) the incorporation of a deep learning-based low-light enhancement model directly into the colourisation pipeline to ensure consistent performance under poor illumination, and (iii) the creation of an optimised, hardware-agnostic software package capable of colourising the entire point cloud in real time.

Together, these contributions move beyond incremental improvements and form the basis of a new, integrated system architecture. Unlike previous approaches that rely on single-camera coverage, require fiducial targets, or fail under reduced illumination, the proposed framework delivers a complete solution that unifies calibration, colour correction, low-light visual enhancement, and optimisation into one workflow. This integration not only ensures high-quality colourised outputs under diverse conditions but also provides a scalable foundation for building compact, field-ready LiDAR–camera monitoring devices that can be adapted to challenging real-world environments such as underground mines, autonomous navigation, and night-time monitoring.

## 2. Materials and Methods

*2.1. Hardware Components*

The hardware setup, chosen to demonstrate the operational concept, consists of four main components: a Velodyne Puck Hi-Res LiDAR sensor, four cameras, an Intel NUC computer that serves as the processing unit, and the Anker 737 battery that provides power to the system.

The Velodyne Puck Hi-Res is a 16-channel mechanical LiDAR sensor capable of creating three-dimensional point clouds at distances of up to 100 meters. It offers a full 360-degree horizontal field of view (HFOV) and a 20-degree vertical field of view (VFOV), operating at a rotational speed of 10 Hz [27]. To extract colour information, the system integrates multiple cameras sourced from different manufacturers (Logitech, ZIQIAN, GLISTA, Uniarch). This diversity highlights the versatility, scalability, and adaptability of the proposed algorithm, demonstrating that it is not limited to any specific camera make or model. Each of these cameras can produce high-definition (HD) images and features a horizontal FOV within the range of 90°–100° and a vertical FOV within 60°–80°. The number of cameras was determined based on their collective fields of view to ensure full coverage of the LiDAR's scanning range.

The LiDAR and cameras were interfaced with an Intel-NUC computer equipped with a Core i5-5250U CPU running the Ubuntu 20.04 operating system. The system is powered by the Anker 737 Power Bank, which supplies an average of 36W and provides ≈ 3 hours of runtime.

*2.2. Hardware Assembly*

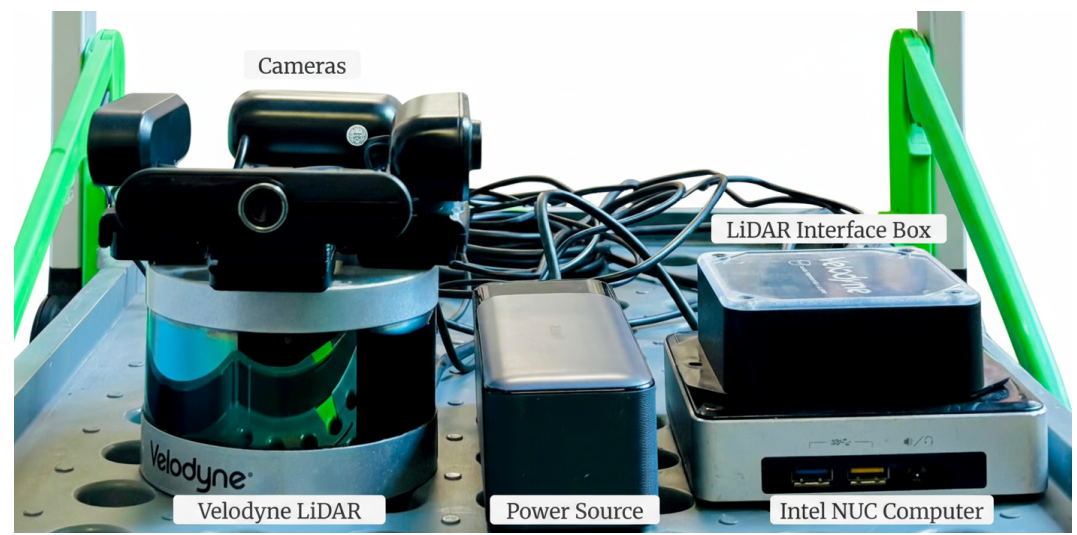

**Figure 1.** Prototype hardware setup on a trolley: Four cameras mounted around the Velodyne Puck Hi-Res LiDAR, LiDAR interface box, Intel NUC for data processing and Anker 737 Power Bank for power supply.

The design, as shown in Figure 01, was simple yet effective. The four cameras were carefully mounted on the top surface of the LiDAR, each facing outwards and positioned 90° apart, providing a combined field of view (FOV) that nearly covers the entire LiDAR scanning area. Due to the physical arrangement of the cameras, positioned at different vantage points rather than a shared optical center, angular blind spots appeared between adjacent views. These gaps vary slightly depending on each camera's individual FOV, resulting in an uneven overlap distribution. To quantify this, the minimum radius at which adjacent cameras offer continuous coverage was calculated by intersecting their inner FOV boundaries, as illustrated in Figure 2. For two cameras separated by an azimuthal angle $\alpha$, with offsets $d_1$ and $d_2$ from the LiDAR centre and half-FOVs $\theta_1$ and $\theta_2$, the intersection radius is defined as,

$$R = |p_1 + \frac{((p_2-p_1) \times u_2)}{u_1 \times u_2} u_1| \tag{1}$$

Where $p_1 = (d_1, 0)$, $p_2 = (d_2 \cos \alpha, d_2 \sin \alpha)$, $u_1 = (\cos \theta_1, \sin \theta_1)$, $u_2 = (\cos(\alpha - \theta_2), \sin(\alpha - \theta_2))$ and $a \times b = a_x b_y - a_y b_x$ is the 2D cross product. As shown in Figure 2b, evaluating $R$ across all adjacent camera pairs gives a minimum radius of approximately 1.31 m, beyond which complete 360° coverage is achieved.

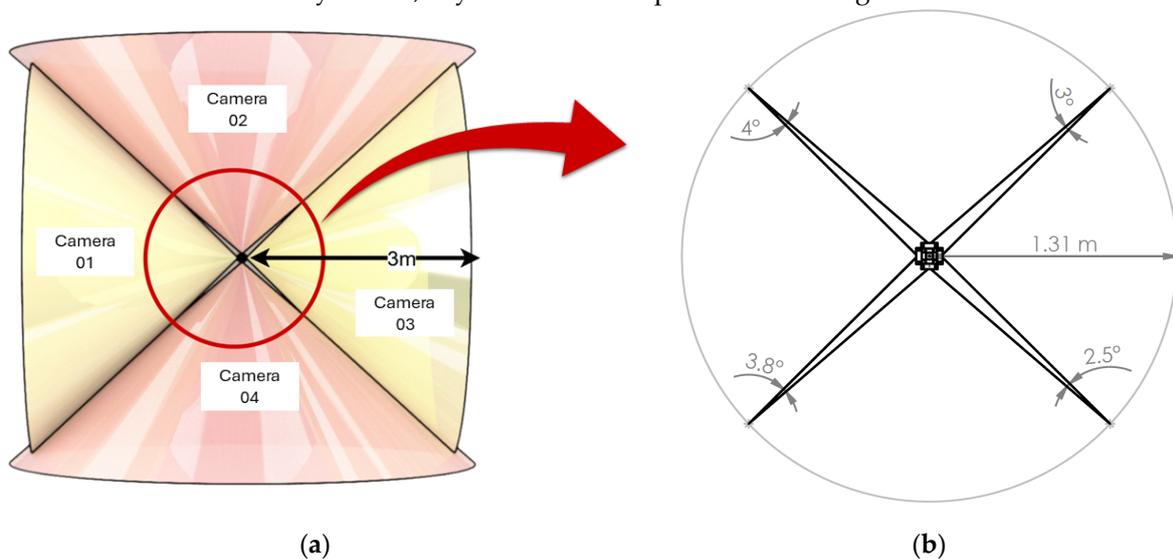

(a) (b)

**Figure 2.** Camera coverage analysis and blind spot identification: (**a**) schematic of camera coverage within a 3-meter radius where yellow regions indicate cameras with a lower FOV, and red regions correspond to cameras with a higher FOV. (**b**) Top view 2D diagram highlighting blind spot regions with four narrow angular gaps, labelled with their respective angular values.

*2.3. Software Architecture*

Two distinct software packages have been developed for calibration and data processing. A neural network model to enhance low-light images was also designed and trained. These packages were written in C++ and Python, using PCL, OpenCV and PyTorch dependencies.

The calibration package performs three key functions. First, it determines each camera's intrinsic parameters to correct image distortions and enable accurate projection of points from the camera coordinate frame onto the 2D image plane. Second, it calculates colour correction coefficients to minimise visual inconsistencies across cameras, ensuring uniform appearance in the final output. Third, it establishes the geometric transformation between the LiDAR and each camera, aligning their coordinate frames for accurate spatial fusion.

Data acquisition and processing were managed through the Robot Operating System (ROS) Noetic, an open-source middleware framework. The pipeline was structured as modular nodes, each with specific responsibilities: one subscribed to the raw LiDAR and camera streams for data acquisition, another synchronised the inputs by aligning timestamps across sensors, a third applied colour correction to minimise inter-camera variations, a fourth enhanced the synchronised images under low-light conditions using the trained neural model, and a final node fused the corrected and enhanced images with LiDAR data to generate and publish high-fidelity, colourised point clouds in real time.

*2.4. Data Capture*

The data capture process for the initial calibration of the hardware and final testing phases is outlined as follows:

2.4.1. Data for camera intrinsic calibration

A checkerboard pattern consisting of 10x7 squares, each measuring 40 mm, was printed on sturdy cardboard. This cardboard pattern was mounted on a wooden frame with a handle for easy movement during the data capture. The hardware setup was placed on a flat surface, and calibration was performed for one camera at a time. The checkerboard pattern was systematically moved in front of each camera, ensuring the entire board was visible within the field of view. The images were captured and used to calculate the camera intrinsic parameters of each camera.

2.4.2. Data for LiDAR Camera extrinsic calibration

The hardware prototype was placed inside a room, and LiDAR and camera feeds were captured. The calibration pipeline was fully automated, making it independent of the objects present in the environment. The setup was moved and rotated by slight distances and angles from a stationary position. All the camera images and the LiDAR frame from the initial stationary position were recorded and used for the extrinsic calibration process.

2.4.3. Data for colour correction calibration

A standard colour checker chart with 24 colour patches was used as the reference. It provides known and consistent colour properties. The chart was displayed under identical lighting and captured by each camera.

2.4.4. Data for outcome evaluation

The calibrated device was placed on a platform, and two 25-second datasets were collected at an intersection of four corridors in the UNSW K15 building. During data capture, the LiDAR platform was gently tilted around its vertical axis to increase the vertical FOV but was not rotated. The data was recorded under both dark and well-lit conditions. Details of these datasets are provided in Table 1.

**Table 1.** Overview of datasets collected in different lighting conditions.

| Dataset Name | Duration (s) | Data Rate (Hz) | | Light Source | Light intensity (lx) |
|---|---|---|---|---|---|
| | | Cameras | Lidar | | |
| LIGHT dataset | 25 | 30 | 10 | Fluorescent lights | 480 |
| DARK dataset | 25 | 30 | 10 | Minimal ambient light | 0.5 |

These two datasets were used to evaluate the system's performance under different lighting conditions. The LIGHT dataset served as the reference in bright light settings, while the DARK dataset was used to assess the effectiveness of the proposed method in low-light environments.

*2.5. Method*

The methodology, illustrated in Figure 03, began with a one-time calibration process, setting the stage for subsequent data fusion processes. Camera calibration took place by displaying a checkerboard pattern in front of the camera and capturing images from varying orientations. The developed algorithm then identified the corners of the checkerboard squares in these images and estimated the camera's intrinsic parameters.

The next phase was to determine the geometric transformation between the LiDAR and camera coordinate frames. An automatic, targetless method was developed. First, a 3D scene was reconstructed using images captured by the selected camera. The resulting 3D reconstruction and the corresponding LiDAR frame were filtered to remove unwanted points, including ceiling, ground points, and walls. The filtered point clouds were then clustered and classified into distinct objects. These objects from the image-based reconstruction and the LiDAR frame were registered, and the alignments were optimised to calculate the rigid transformation between the LiDAR and the camera. This geometric relationship was then used to compute the transformations for the other three fixed cameras.

There was a noticeable contrast and brightness difference among the images from each camera. To address this issue, all the cameras were calibrated to match the colours of a colour reference image. This completed the initial calibration process, preparing the system for operation.

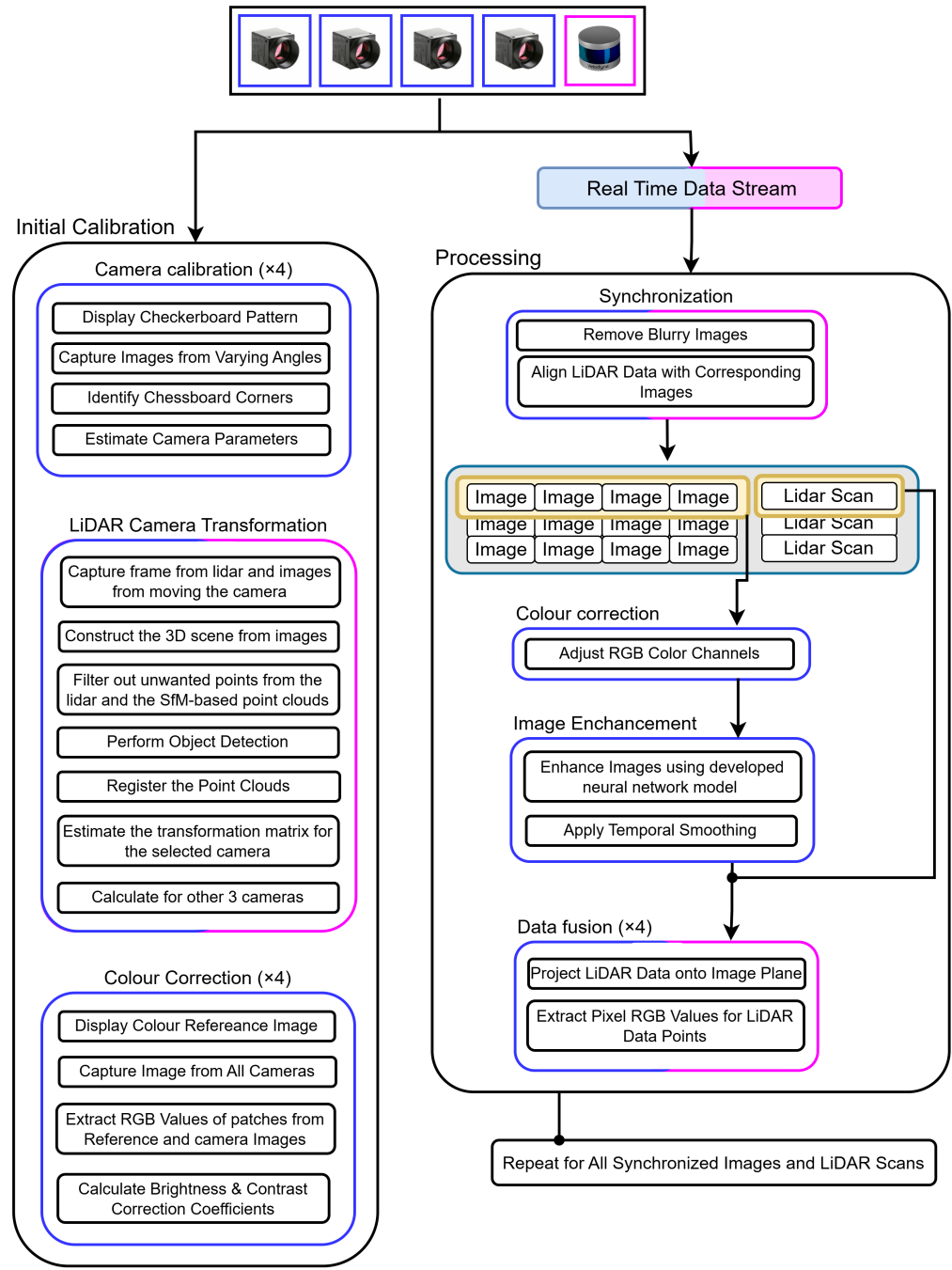

**Figure 3.** Complete workflow of the proposed method. The method has two phases: Initial Calibration (left) and Processing (right). Calibration data provides proper alignment of cameras and LiDAR, which is then used in real-time to process incoming data for accurate data fusion.

The data acquisition pipeline used ROS packages to collect real-time data, which then underwent a series of processing steps. Blurry images were first filtered out. Each LiDAR frame was then paired with the four temporally closest focused camera frames. These selected images were then adjusted using brightness and contrast correction factors specific to each camera. A neural network model developed to improve image quality in low-light conditions further enhanced these colour-adjusted images. This enhancement was applied only when image brightness fell below a specified threshold. Temporal smoothing was then applied to ensure consistency across frames.

Following these preprocessing steps, the data fusion process began. LiDAR points were projected onto the camera plane using the LiDAR-to-camera transformation matrix, and the colour values of the projected points were extracted by interpolating the surrounding image pixels. This process was repeated for all four cameras with their respective LiDAR frames, and any duplicate points caused by overlapping fields of view were removed. The resulting colourised point cloud data was stored for further analysis.

2.5.1. Camera Intrinsic Calibration

Image distortion affects the alignment between the camera and LiDAR by altering the geometric representation of the scene, leading to inaccuracies in point projection. To correct this, cameras were calibrated. It involves determining the camera matrix, including the focal lengths and optical centers, and calculating distortion coefficients to correct radial and tangential distortions. Additionally, the skew coefficient is assessed to correct any misalignment between the x and y pixel axes. This calibration process was executed using the well-established functions in the OpenCV library [28-30]. The calculated camera parameters were then applied to correct the images, resulting in geometrically accurate representations.

2.5.2. LiDAR-Camera Extrinsic Calibration

LiDAR camera extrinsic calibration involves determining the relative rotation and translation matrix between the LiDAR and camera coordinate frames. The proposed calibration process was inspired by the method proposed by Yoon et al. [25]. It was fully automatic and didn't rely on specific geometric objects in the scene. The process has 03 main steps: creating a 3D point cloud from images, clustering and classifying the point cloud, and performing registration. An overview of the process is shown in Figure 4.

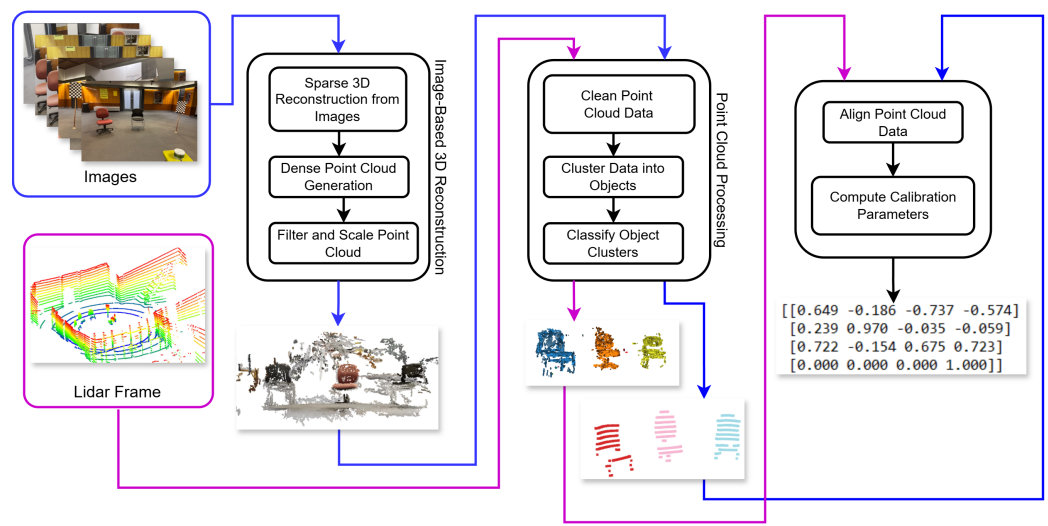

**Figure 4.** Overview of the proposed LiDAR camera calibration process

A set of images from one camera and corresponding LiDAR data was taken. These images were processed using the Structure from Motion (SfM) algorithm to create a 3D reconstruction. The SfM process started with feature detection. The Scale-Invariant Feature Transform (SIFT) algorithm was used to detect key points and create descriptors of these points in each image. These descriptors were then matched across images, linking keypoints that represent the same physical features observed from multiple viewpoints [31]. From these 2D–2D correspondences $(x_i, x'_i)$, where $x_i$ and $x'_i$ are the homogeneous coordinates of the matched keypoints in two images, the fundamental matrix $F$ was estimated such that it satisfies the epipolar constraint:

$$x'^T_i F x_i = 0 \qquad (2)$$

The fundamental matrix captures the geometric relationship between two uncalibrated views, ensuring that a point in one image must lie on a corresponding line in the other. With the camera intrinsic matrix $K$, Essential matrix $E$ was then computed from the fundamental matrix as:

$$E = K^T F K \quad (3)$$

$$E = [t]_\times R \quad (4)$$

Where $[t]_\times$ is the skew-symmetric matrix of the translation vector $t$ between two cameras, and $R$ is the relative rotation. Decomposition of $E$ provides the solution for camera poses.

With the known camera poses and the matched key points, triangulation was applied to recover the 3D coordinates of each feature point. For each correspondence, the triangulated 3D point $X_i$ was obtained by minimising the reprojection error in both camera views:

$$X_i = \arg\min_X (|x_i - K[R|t]X|^2 + |x'_i - K[R'|t']X|^2) \quad (5)$$

This produced a sparse 3D point cloud, which was further refined using bundle adjustment. Bundle adjustment jointly optimised the estimated camera poses and 3D point positions by minimising the global reprojection error across all views. Here $k$ denotes the camera view index, $i$ is the 3D point index, and $x_{i,k}$ represents the observation of the 3D point $X_i$ in view $k$. The optimisation is formulated as:

$$\min_{R_k, t_k, X_i} \sum_{k,i} |x_{i,k} - K[R_k\ t_k]X_i|^2 \quad (6)$$

The refined sparse point cloud was then densified using the Multi-View Stereo algorithm (MVS) [32]. The resulting point cloud was filtered using the statistical outlier filter to remove noisy measurements [33]. The final dense 3D point cloud, along with the estimated camera poses and the corresponding 2D image data, was saved.

Prior to registration, both the SfM and LiDAR point clouds were preprocessed to isolate objects of interest. The ceiling points were removed using a height threshold, and the ground points were removed using the Simple Morphological Filter (SMRF) algorithm [34]. The walls were removed by calculating the surface normals and identifying planes perpendicular to the ground plane. The remaining points were segmented into distinct object clusters using the DBSCAN clustering algorithm. These preprocessing steps ensured a clear separation of objects into well-defined clusters. The clusters were classified into semantic categories using the PointNet++ algorithm [35], trained on the S3DIS dataset [36].

Since the SfM reconstruction does not inherently preserve metric scale, it was necessary to rescale it to match the LiDAR measurements. The scale factor was calculated by comparing the average centroid distances between object clusters in the SfM reconstruction and those in the LiDAR point cloud. In this step, $c(\cdot)$ denotes the centroid of an object cluster, $V_L$ and $V_{SfM}$ are the LiDAR and SfM object point clouds, respectively, and $p, q$ are indices of object clusters, with averaging performed over all pairs of clusters. The scale factor was computed as:

$$s = \frac{\text{avg}_{p,q} |c(V_L^p) - c(V_L^q)|}{\text{avg}_{p,q} |c(V_{SfM}^p) - c(V_{SfM}^q)|} \quad (7)$$

Multiplying the SfM clusters by the factor s aligned the reconstruction with the metric scale defined by the LiDAR measurements.

Following scaling, the corresponding clusters from the LiDAR frame and the SfM point cloud were registered using the Iterative Closest Point (ICP) algorithm. RANSAC-based ICP was first applied to obtain a robust initial alignment without requiring an initial transformation guess. The results were then refined using the point-to-plane ICP. This registration was performed separately for each cluster, and the resulting transformations were jointly optimised in a global step that considered all clusters together, producing a consistent alignment between the two modalities.

The optimised transformation was converted from the SfM coordinate frame to the camera coordinate frame using the camera poses derived from the essential matrix (Eq. 4). This resulted in the required camera–LiDAR extrinsic calibration matrix. All remaining cameras were calibrated simultaneously using their respective image sets in the same SfM reconstruction. The scaling and registration procedure used for the first camera was extended jointly, resulting in extrinsic matrices for all cameras in a single step. The entire pipeline was implemented in Python and automated as part of the calibration pipeline. The resulting matrices were validated against the mounting orientations specified in the 3D design drawing.

2.5.3. Colour Correction Calibration

The colour profiles of the cameras varied significantly even within the same camera model. Merging them negatively affects the results, leading to a lack of colour uniformity within the point cloud frames. To achieve a uniform colour profile across cameras, initial tests were conducted using histogram equalisation [37] and normalisation methods [38]. These methods, although commonly used for enhancing the contrast of images, proved ineffective for our requirement. The main reason was that these techniques depended only on the colour intensity distribution of an isolated image and did not consider a standard or specific colour palette of the reference image.

A simple yet effective algorithm was developed. A camera colour correction card with known colour profiles was used. It served as a reference. The objective was to match the colour schemes of images from the four cameras to that of the reference image, focusing on one image at a time. This alignment can be achieved as described by,

$$\text{Colour-corrected image} = \text{contrast coefficient} \times \text{reference image} + \text{brightness coefficient} \quad (8)$$

The pixel-to-pixel matching method was used, where the RGB values of each area within the 24 colour patches of the reference image were compared with their absolute values (Figure 5a). Linear regression analysis was performed for each colour channel (R, G, B). The slope and intercept of the regression line were used as the contrast and brightness coefficients, respectively, for that channel in a specific camera, as shown in Figure 5b. Each camera, therefore, produced six calibration values—two for each colour channel. These coefficients were recorded and later applied during the colour correction phase. This calibration procedure was repeated for the images captured by the other three cameras, ensuring consistent colour correction across all cameras.

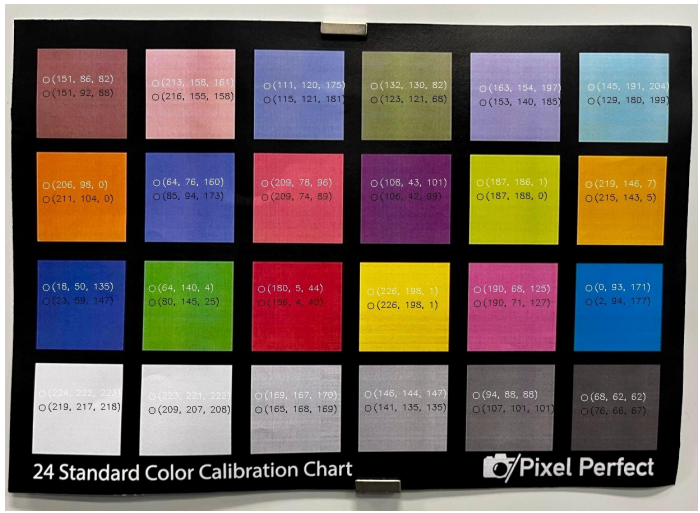 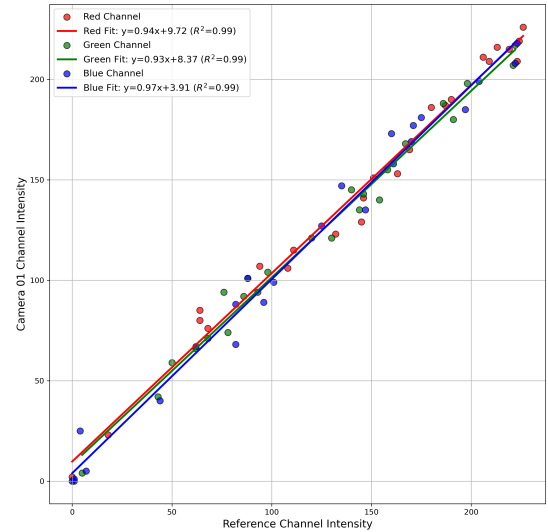

(**a**) (**b**)

**Figure 5.** Colour correction calibration process (**a**) Camera colour correction card. White text: RGB values from camera 01, Black text: Reference RGB components (**b**) Linear regression for colour channels with the fitted equations and the coefficient of determination ($R^2$) values indicated in the legends.

2.5.4 Data Synchronisation

The LiDAR and cameras operate at different frequencies: LiDAR provides a point cloud data frame at 10Hz, equating to one frame every 100 milliseconds, while the cameras capture at a rate of 30 FPS. This frequency mismatch necessitates a precise calibration to align the spatial data from the LiDAR with the corresponding visual data from the cameras to the same temporal point.

During the process, the blurry images were filtered out using the variance of the Laplacian method. This technique estimates image sharpness by applying the Laplacian operator, which detects edges based on second-order derivatives. The variance of the resulting values serves as a measure of blurriness, with lower variance indicating a blurrier image [6,39]. If the variance fell below a predefined threshold of 150, calculated based on the visual observations of images from Section 2.4.4, the image was classified as blurry.

For synchronisation, camera and LiDAR timestamps were taken from the common ROS time base. Let $t_k^C$ denote the timestamp of the k-th camera frame and $t_m^L$ the timestamp of a LiDAR frame. For each LiDAR frame, the camera frame minimising $|t_k^C - t_m^L|$ was identified. If the difference was less than 16 ms (half of the 33 ms camera interval), that camera frame was selected. In this way, one sharp image from each camera was paired with every LiDAR frame.

2.5.5. Image Enhancement

The U-Net architecture-based model [40] was developed and trained for low-light image enhancement. The neural network architecture, shown in Figure 6, was carefully designed to handle images of any input dimensions and output enhanced images with the same dimensions. The model consisted of two main components:
- Encoder (Contracting Path): The encoder reduces the spatial dimensions of the image through layers of convolutions and pooling operations. The feature extraction head of the pre-trained VGG19 model (trained on ImageNet) was used as the encoder.
- Decoder (Expansive Path): Mirroring the encoder, the decoder gradually reconstructed the target output from the feature set learned by the VGG19 encoder.

The model is further improved by incorporating attention layers. These layers serve as skip connections, effectively linking the encoding and decoding segments of the network.
- Spatial Attention: Helped the model focus on specific regions within the image that were most affected by low-light conditions, such as excessively dark areas or shadowed regions.
- Channel Attention: Enabled the model to prioritise features or channels carrying critical information about edges, textures, or colour, which were essential for restoring the image.

The Charbonnier loss function [41] was used to quantify the difference between the model's predictions and the ground truth images. This function was designed to be less sensitive to outliers than other loss functions, which helps in getting clearer and more accurate image enhancement results. The Adam optimiser was used to guide the neural network parameters, minimising the Charbonnier loss during the training process.

The SID (See-in-the-Dark) [42] and ExDark [43] were used, as they provided sufficient data for training. The dataset was randomly split into training, validation, and testing sets, with 70% allocated for training, 15% for validation, and 15% for testing. All images were normalised to scale pixel intensity values to the 0–1 range. Padding was applied to the images, adding extra pixels around the edges to maintain compatibility with the network's pooling and up-sampling layers, ensuring that the final output matched the input dimensions.

Hyperparameters were established, including a learning rate of 1e-4, a batch size of 16, and 100 training epochs. The normalised and padded low-light images were fed into the network, and Weights were iteratively adjusted based on the calculated loss using the Adam optimiser. After training, the model was tested on the testing set containing unseen low-light images. The trained model weights were then saved.

These saved weights were used during the real-time processing pipeline to enhance incoming colour corrected camera images with a resolution of 1280x720 pixels. A predefined brightness threshold of 0.12 (on a normalised scale from 0 to 1) was applied to determine whether enhancement was required. This threshold was selected based on empirical analysis of the training and testing datasets, along with visual inspection of images from the DARK dataset. When the mean brightness of an image fell below this threshold, it was classified as too dark for reliable colour extraction. In such cases, the image was passed through the enhancement model, which produced an output image of the same input resolution with improved contrast and recovered semantic and structural details that would otherwise have remained indistinguishable.

The model produced enhanced outputs with a consistent colour appearance across all four camera views, owing to the prior colour correction process. However, in some instances, significant contrast changes were observed among successive frames from the camera feeds. This was expected since the model was designed to enhance individual images, not sequences. Temporal smoothing was applied to address this, which blends information from the current and adjacent frames to yield a smoother output [44,45]. It ensured that individual images appeared visually appealing and the entire sequence maintained visual coherence.

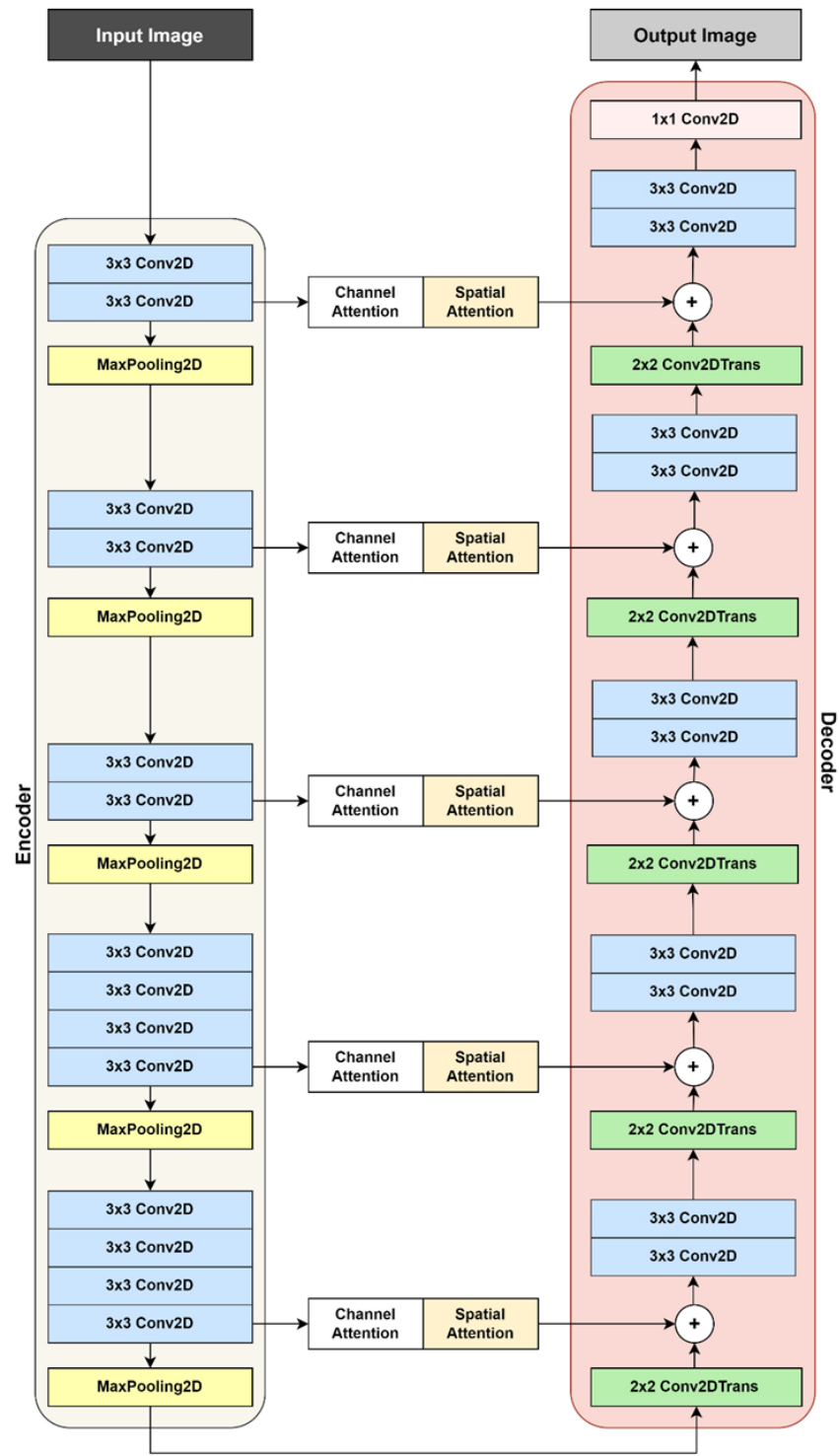

**Figure 6.** Image enhancement model architecture

2.5.6. Camera and LiDAR Data Fusion

The final step in the fusion process, summarised in Algorithm 1, combined synchronised LiDAR frames with colour-corrected images from four cameras. For each camera, the raw image was first undistorted using its intrinsic matrix and distortion coefficients. The LiDAR point cloud was transformed into the camera's coordinate frame using the precomputed extrinsic transformation matrix. Points with negative depth (z < 0) were discarded to avoid invalid projections.

**Algorithm 1** LiDAR-Camera Fusion

1: **Input:** LiDAR point cloud (300,000 points), 4 camera images; for each camera c: intrinsic parameters **K**c, distortion coefficients **D**c, and camera-LiDAR transformation (**R**c, **T**c); Method ∈ {loop-based, vectorised}
2: **Output:** Colourised LiDAR point cloud
3: Define image bounds:
4:     $w_{min}$ = 0, $w_{max}$ = image width (1280)
5:     $h_{min}$ = 0, $h_{max}$ = image height (720)
6: **for** each LiDAR frame **do**
7:     **for** each camera c (4 cameras) **do**
8:         undistorted_image ← **Du**(image$_c$, **K**c, **D**c)     ▷ Apply distortion correction
9:         points_cam$_c$ ← **R**c · points + **T**c     ▷ Transform LiDAR points to camera
10:         **if** points_cam$_c$(z) < 0 **then**
11:             Remove points     ▷ Filter out points with negative depth
12:         **end if**
13:         image_projected$_c$ ← **K**c · points_cam$_c$     ▷ Project points using
14:     **end for**
15:     **if** Method = loop-based **then**
16:         **// Loop-based colour assignment**
17:         **for** each camera c (4 cameras) **do**
18:             **for** each projected point (u, v) in image_projected$_c$ **do**
19:                 **if** u < $w_{min}$ or u ≥ $w_{max}$ or v < $h_{min}$ or v ≥ $h_{max}$ **then**
20:                     Assign default colour (0, 0, 0)
21:                 **else**
22:                     Assign colour by interpolating over a 3 × 3 neighbouring grid
23:                 **end if**
24:             **end for**
25:         **end for**
26:     **end if**
27:     **if** Method = vectorised **then**
28:         **// Vectorised colour assignment**
29:         **for** each camera c (4 cameras) **do**
30:             Create valid mask:
31:                 valid_mask$_c$ ← (image_projected$_c$(x) ≥ $w_{min}$) ∧ (image_projected$_c$(x) < $w_{max}$) ∧ (image_projected$_c$(y) ≥ $h_{min}$) ∧ (image_projected$_c$(y) < $h_{max}$)
32:             Initialise colour_map$_c$ with default colour (0, 0, 0)
33:             Assign colours:
34:                 colour_map$_c$(valid_mask$_c$) ← undistorted_image$_c$(image_projected$_c$(valid_mask$_c$))
35:         **end for**
36:     **end if**
37: **end for**
38: **Return:** Colourised LiDAR point cloud

The remaining LiDAR points in the camera coordinate frame were then projected onto the image plane using the camera's intrinsic matrix. The resulting pixel coordinates were checked against the image boundaries. If the projection fell within the valid image region, colour values were assigned to each point by averaging over a 3×3 neighbourhood of surrounding pixels. Points outside the image bounds were assigned a default RGB value of (0, 0, 0).

This procedure was repeated for all four camera views, enabling full 360° coverage and maximising the number of LiDAR points enriched with colour information. The resulting colourised point cloud, containing point indices and corresponding RGB values, was stored for downstream processing.

The colourisation pipeline was first implemented in Python using a point-wise loop, where each LiDAR point was transformed, projected, and assigned colour individually; although simple, this approach was computationally expensive because identical matrix operations were repeated for every point, making the processing of dense point clouds inefficient. To improve efficiency, the pipeline was re-implemented in C++ using a vectorised, matrix-based design in which the entire LiDAR point set was represented as a matrix, allowing transformation, projection, and boundary checks to be performed in batch through optimised linear algebra routines, as outlined in Algorithm 1.

## 3. Results

### 3.1. Camera Intrinsic Calibration

The camera calibration was conducted for the 04 cameras. Camera internal parameters were estimated. Distortion parameters were also determined, and adjustments to the perspective were made, resulting in the checkerboard lines appearing more parallel and straight. The calibration process additionally estimated the extrinsic parameters, defining the position and orientation of each camera relative to the scene. These parameters were used to compute the reprojection error, which quantifies the accuracy of the calibration. An average reprojection error of 0.35 pixels was recorded, indicating robust calibration.

### 3.2. LiDAR-Camera Extrinsic Calibration

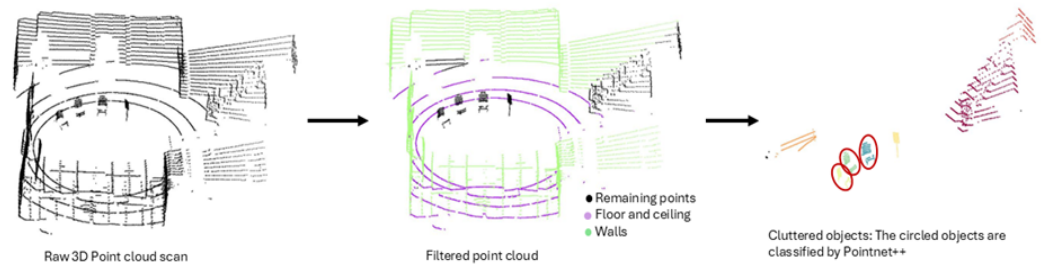

**Figure 7.** 3D point cloud processing pipeline: From raw point cloud to filtered data, clustering and classification

Figure 7 presents the extrinsic calibration results for Camera 01 and the LiDAR sensor. The ceiling, floor, and vertical walls were removed. This resulted in a cleaner point cloud containing only points related to the objects of interest. This step prevents the merging of multiple objects into a single cluster and avoids floor and wall points being incorrectly assigned to object clusters. Clustering effectively separated the objects, and PointNet++ classified them into distinct categories. A similar process was applied to the point cloud generated from images, yielding satisfactory results.

The registration of the remaining points from these two-point clouds was conducted, with the results shown in Figure 8(a). The chairs in the scene were correctly aligned with the LiDAR points (green), and the overlay was accurate, with precise edge matching. Gaps in the SfM reconstruction occurred because the images were taken from slightly varying orientations without covering all possible perspectives. Still, ~84% of LiDAR points from the classified objects found their counterparts in the corresponding objects detected in the SfM reconstruction. The percentage of matching points for each individual object, referred to as the correspondence score, is shown in Figure 8(a). This correspondence allowed for the accurate computation of the LiDAR-camera transformation matrix.

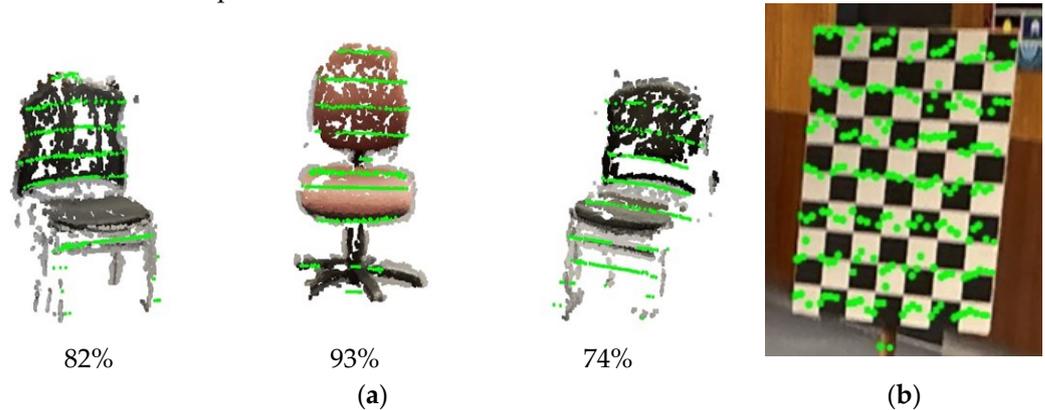

82%   93%   74%

(**a**)        (**b**)

**Figure 9.** (**a**) Results of the point cloud registration process: Alignment of the LiDAR points with the SfM-generated point cloud. The correspondence score for each object is indicated below the respective object, representing the percentage of LiDAR points that found matches in the SfM reconstruction. (**b**) Validation of the transformation matrix by projecting LiDAR points onto a checkerboard plane.

To verify the calibration, LiDAR data and corresponding camera images containing the checkerboard (previously used for intrinsic calibration) were captured. Checkerboard points in the LiDAR frame were manually extracted, and their corresponding image coordinates were identified. The extracted LiDAR points were then projected onto the image using the inverse of the LiDAR–camera extrinsic transformation matrix. The results are shown in Figure 8(b). The precise alignment, with average translation and rotation errors of 2.91 mm and 0.31°, respectively, and a reprojection error of 1.71 pixels, confirms the high accuracy of the registration and calibration process.

The estimated LiDAR–camera extrinsic matrices were further validated by calculating the relative position of the camera with respect to the LiDAR and comparing it with the mounting orientations specified in the 3D rig drawings. The recovered camera position differed from the design geometry by an average of 2.1 mm in translation, and the orientation differed by less than 0.4°. These deviations are within the acceptable range defined by the mechanical tolerances of the mounting assembly and the alignment precision achievable during installation. This validation confirmed that the estimated calibration was consistent with the intended geometry of the mountings specified in the drawings.

*3.3. Colour Correction Calibration*

The calibration process with the reference image and the image captured by camera 01 is shown in Figure 5. The coefficient of determination ($R^2$) value of 0.99 for all three channels (Red, Green, and Blue) indicates a near-perfect fit, showing that the data aligns closely to a straight line. A similar relationship was observed across the other three cameras, where all the channels achieved $R^2$ value above 0.98. These results demonstrate that the images from each camera are closely matched to the reference pixel intensity values, ensuring consistent colour mapping across the camera set.

*3.4. Image Enhancement*

The model was evaluated using the testing dataset, recording a Peak Signal-to-Noise Ratio (PSNR) value of 26.634 and a Structural Similarity Index Measure (SSIM) value of 0.792. The model is quantised, and an FPS rate of 50+ was achieved with a single GPU, which is more than sufficient for this application, as the cameras operate at a maximum output rate of 30 FPS.

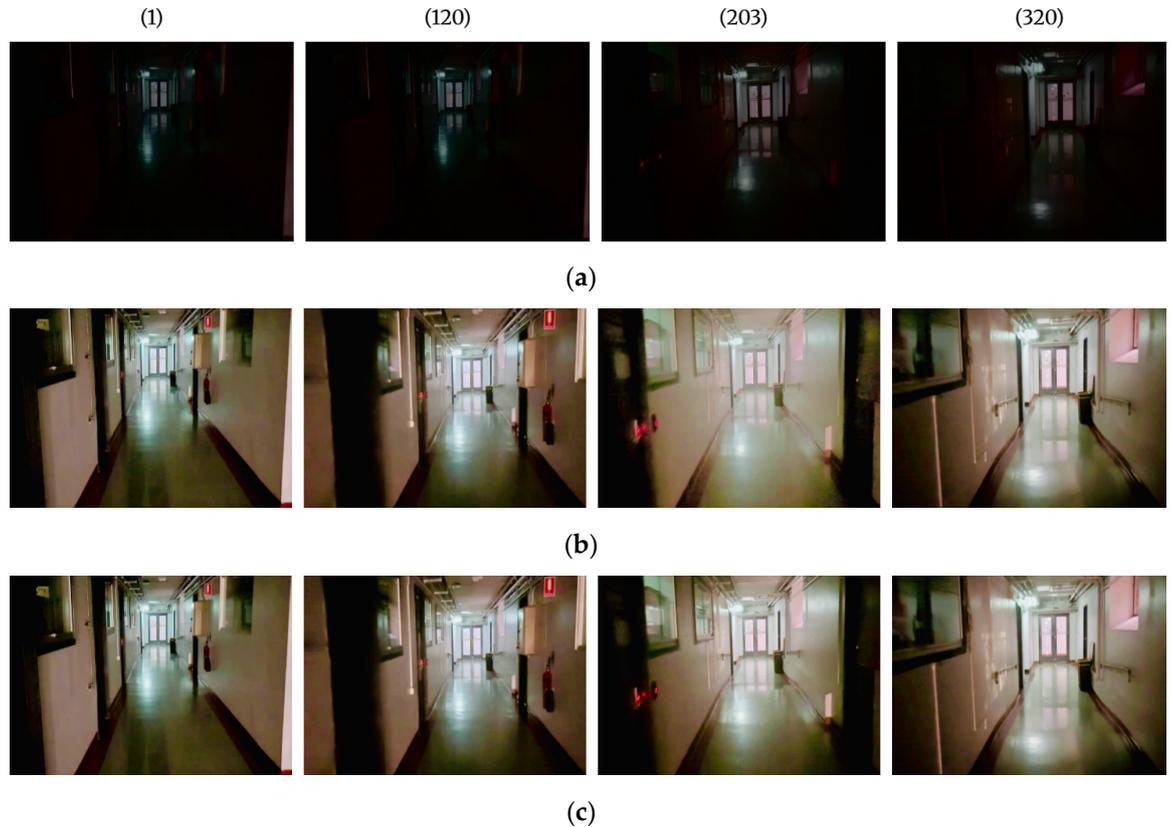

**Figure 9**. Results of low-light image enhancement on a sequence of images taken at a light level of 1 lux: displayed for frames 1, 120, 203, and 320. (**a**) Raw input image set, (**b**) Images after low-light enhancement process, (**c**) Output after temporal smoothing, with nearly uniform colours across the frames.

The results presented in Figure 9 demonstrate the significant impact of the low-light image enhancement model on enhancing dark images. In the raw frames, scene features were poorly visible, whereas the enhanced images show a substantial enhancement in feature visibility, with most structures becoming clearly distinguishable. Despite these improvements, the enhanced images in frame 203 exhibit fluctuations in contrast and brightness. The application of temporal smoothing reduces the overexposed effects in the images, achieving a more balanced and natural appearance across the frames.

*3.5. Camera and LiDAR Data Fusion (Final Output)*

The Python implementation of the loop-based data fusion algorithm resulted in a processing frequency of 1.2 Hz, which is significantly lower than the 10 Hz LiDAR input rate. The C++ implementation improved execution speed by 4×, achieving ~ 5 Hz. This improvement was primarily due to C++ being more efficient in handling ROS messages and its direct integration with OpenCV and PCL libraries, which reduces overhead associated with Python's interpreted execution and memory management. By replacing the loop-based implementation with a vectorised approach that eliminates explicit loops, the system achieved a processing rate of 10 Hz, matching the LiDAR input rate and enabling real-time performance. This vectorised method processes multiple data points simultaneously, significantly reducing redundant operations, minimising memory access delays and optimising cache utilisation, leading to a substantial improvement in execution speed.

The fused data reveal that an average of 99.9% of points from the LiDAR frames have been successfully colourised, achieving an almost complete 360-degree HFOV with full LiDAR VFOV. The remaining 0.1% were data points that fell within the blind spots of the device, a limitation of the four-camera setup. This level of colourisation is a notable achievement, as other algorithms typically colourise only a limited portion of the LiDAR scan.

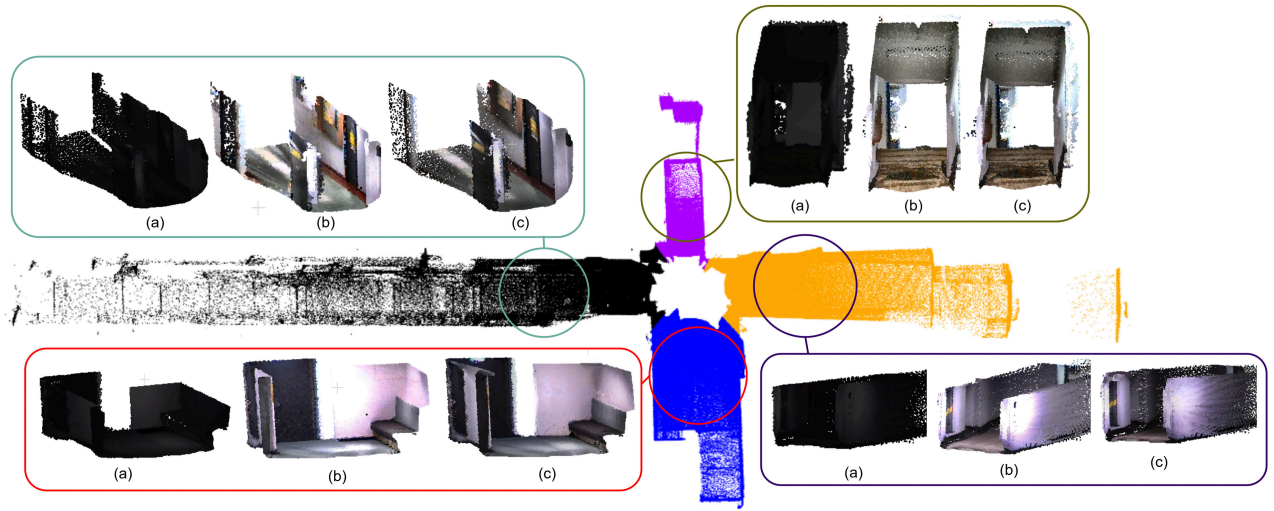

**Figure 10.** SLAM-based 3D reconstruction with colour-coded camera contributions and patch-wise visual comparison of (a) and (b) correspond to the DARK and LIGHT datasets without low-light enhancement, and (c) corresponds to the DARK dataset with low-light enhancement applied.

The two datasets (referenced in section 2.4.3) provide an appropriate platform to compare the final outputs coming from 02 different pathways. The LIGHT dataset allows the evaluation of system performance under well-lit conditions, where images are vibrant and rich in light.

The captured data was first processed using the LOAM (LiDAR Odometry and Mapping) algorithm [46] to generate the geometric map from the LiDAR point clouds. The colour values obtained through the calibration process were then added as a scalar field to the point cloud, producing the final colourised map shown in Figure 10. Reconstructions from the LIGHT dataset appears to be bright and colourful. In contrast, the original output from the DARK dataset—without enhancement—is nearly black, lacking visible textures. However, when low-light enhancement is applied, the results from the DARK dataset closely resemble those from the LIGHT dataset. Despite originating from near pitch-black conditions, the enhanced reconstruction maintains lighter colours and visual detail. This transformation—from initially dark and texture less point clouds to vibrant, detailed reconstructions—highlights the effectiveness of the proposed algorithm in overcoming challenging lighting conditions and achieving consistent reconstruction quality.

## 4. Discussion

The results of the study demonstrate the versatility of the LiDAR point cloud colourisation approach across a wide range of lighting conditions (Figure 10). By integrating LiDAR with multiple cameras, the method achieves complete 360° coverage, ensuring that nearly every point in the scan is accurately colourised. This significantly improves semantic interpretability compared to existing methods that typically rely on single-camera setups with a limited field of view.

A key differentiator of the proposed method is its robustness in low-light environments, achieved through the incorporation of a dedicated image enhancement module. This module transforms initially dark images into colour-accurate representations that closely resemble those captured under well-lit conditions. Such capability is critical for applications in underground environments and night-time monitoring, where conventional camera–LiDAR fusion systems often fail.

Prior to image enhancement, colour correction is applied to standardise brightness and contrast across all camera feeds. This is a deliberate and effective step, as it provides a consistent baseline that prevents uneven colourisation during the enhancement process. Following enhancement, temporal smoothing is employed to reduce flicker and abrupt transitions, ensuring a visually coherent sequence throughout the entire time-lapse. Together, these correction methods contribute to generating a visually consistent and semantically accurate colourised point cloud, even under varying lighting conditions.

Another key strength lies in the calibration process. Unlike traditional methods that require external targets and multiple acquisitions, the proposed approach estimates the extrinsic parameters for all cameras simultaneously by registering the SfM reconstruction to the LiDAR point cloud. This eliminates the need for manual feature selection and reduces the chance of user-induced errors. Beyond efficiency, this approach is inherently scalable: as more cameras or LiDAR units are added to a system, the same algorithm can be applied to calibrate all sensors at once without significant additional effort. This represents a practical step forward for real-world deployment, especially in environments where frequent recalibration is impractical or where multiple devices need to be calibrated in a uniform and efficient manner.

Concept designs were also tested using a single fisheye camera to simplify the setup. However, once mounted, the fisheye field of view was directed either upward or downward, while the LiDAR scans radially around its horizontal surface. This led to a significant mismatch in coverage: large portions of the fisheye image contained no corresponding LiDAR data, and many LiDAR returns were outside the useful camera field. The resulting reduction in valid correspondences made extrinsic calibration unstable. In contrast, the four-camera configuration provided fields of view that aligned more closely with the LiDAR scan geometry, ensuring sufficient overlap for reliable calibration and consistent colourisation.

Computational performance was a critical focus of this work, as real-time operation is essential for practical deployment in dynamic environments. The initial Python implementation was insufficient for real-time use, but an optimised C++ version dramatically improved processing speed. Further refinement with a vectorised, matrix-based approach enabled the system to operate at the full input rate of the LiDAR sensor. This level of optimisation is particularly important for real-world applications such as underground monitoring and autonomous navigation, where delays in data processing can significantly affect system reliability and usability.

In summary, the proposed framework represents a carefully engineered system where each component — from calibration and image enhancement to data fusion and optimisation — has been deliberately designed and refined to maximise overall performance and provide high-quality colourised outputs under all lighting conditions. By integrating these elements into a unified workflow, the system establishes a robust foundation for a fully integrated, real-time LiDAR–camera monitoring solution capable of continuous environmental scanning in diverse and challenging environments.

## 5. Conclusions

In conclusion, this study presents a comprehensive, hardware-agnostic approach for LiDAR point cloud colourisation using a multi-camera setup enhanced by low-light image processing. The proposed framework addresses critical limitations of existing methods, including limited field of view, inconsistent colour output, and poor performance in low-light environments. By achieving full 360° coverage and delivering consistent, high-quality colourisation across varied lighting conditions, the method significantly improves the semantic interpretability of LiDAR data.

At the core of the system is a robust calibration pipeline that performs accurate intrinsic and extrinsic parameter estimation, applies colour correction across heterogeneous cameras, and automatically computes the LiDAR–camera transformation, removing the need for specialised calibration targets. A neural network-based enhancement module further supports reliable colourisation by improving image quality under low-light conditions. The complete processing pipeline has been optimised for real-time performance through a vectorised, matrix-based implementation, enabling low-latency operation and efficient handling of dense point cloud data. This positions the framework not just as an incremental improvement, but as a novel system architecture that can be directly adapted to diverse hardware platforms and field conditions.

This work lays a strong foundation for the development of an intelligent underground monitoring system tailored to challenging, low-visibility environments. Future work will focus on building a compact, field-ready device for underground mines, integrating all components into a single deployable unit suitable for both mobile and fixed infrastructure. Planned efforts include validation in operational mine settings, long-term performance evaluation, and refinement of the system for durability, reliability, and ease of use in demanding underground conditions.

## 6. Patents

This work has resulted in a patent titled "Enhanced Low-Light Point Cloud Colorisation Method, Device, and Electronic Equipment (增强型低光照的点云着色方法、装置和电子设备)" (Application Number: 202411295521.4), filed by Huari Mining Technology Co., Ltd.

**Author Contributions:** Conceptualisation, P.R.; methodology, P.R., D.P., B.P.; software, P.R.; validation, P.R., D.P., and B.P.; formal analysis, P.R.; investigation, P.R. and D.P.; resources, S.R.; data curation, P.R.; writing—original draft preparation, P.R.; writing—review and editing, all authors; visualisation, P.R.; supervision, B.P. and S.R.; project administration, S.R.; funding acquisition, S.R. All authors have read and agreed to the published version of the manuscript.

**Funding:** The project is supported by the Azure Mining Technology Pty Ltd., an Australia registered and wholly owned subsidiary of China Coal Technology & Engineering Group (CCTEG) under the Project 2022-3-KJHZ005.

**Data Availability Statement:** Not applicable.

**Acknowledgments:** We thank Kanchana Gamage (Technical Officer, School of Minerals and Energy Resources Engineering, UNSW) for his assistance with the experimental activities. We also acknowledge the suggestions received from the project partner, Dr Bingfei Nan and Mr. Chenxi Ye of Tianma Intelligent Control Technology Co., Ltd., Beijing.

**Conflicts of Interest:** The authors declare no conflict of interest.